\documentclass[10pt, a4paper]{article}
\usepackage{lrec-coling2024} 
\pdfoutput=1

\usepackage{todonotes}
\usepackage{xspace}
\usepackage{booktabs}

\newcommand{\Modelsmall}{Albertina 100M PT}
\newcommand{\Modellarge}{Albertina 1.5B PT}

\title{Fostering the Ecosystem of Open Neural Encoders for Portuguese with Albertina PT* Family}

\name{Rodrigo Santos$^{\dag}$, João Rodrigues$^{\dag}$, Luís Gomes$^{\dag}$, João Silva$^{\dag}$, António Branco$^{\dag}$,\\ {\bf \large Henrique Lopes Cardoso$^{\ddag}$, Tomás Freitas Osório$^{\ddag}$, Bernardo Leite$^{\ddag}$}}

\address{
    $^{\dag}$University of Lisbon
    \\
        NLX - Natural Language and Speech Group, Department of Informatics\\
    Faculdade de Ciências,
    Campo Grande, 1749-016 Lisboa, Portugal\\
    \{rsdsantos, jarodrigues, luis.gomes, antonio.branco\}@fc.ul.pt\\
    $^{\ddag}$University of Porto\\
    Faculty of Engineering, Department of Informatics Engineering\\
    Rua Dr.\ Roberto Frias, 4200-465 Porto, Portugal\\
    hlc@fe.up.pt, tomas.s.osorio@gmail.com, bernardo.leite@fe.up.pt\\}

\abstract{
To foster the neural encoding of Portuguese, this paper contributes foundation encoder models that represent an expansion of the still very scarce ecosystem of large language models specifically developed for this language that are fully open, in the sense that they are open source and openly distributed for free under an open license for any purpose, thus including research and commercial usages. Like most languages other than English, Portuguese is low-resourced in terms of these foundational language resources, there being the inaugural 900 million parameter Albertina and 335 million Bertimbau. Taking this couple of models as an inaugural set, we present the extension of the ecosystem of state-of-the-art open encoders for Portuguese with a larger, top performance-driven model with 1.5~billion parameters, and a smaller, efficiency-driven model with 100~million parameters. While achieving this primary goal, further results that are relevant for this ecosystem were obtained as well, namely new datasets for Portuguese based on the SuperGLUE benchmark, which we also distribute openly.
 \\ \newline  }

\begin{document}

\maketitleabstract

%\todo[inline]{Up to eight (8) pages, excluding any number of additional pages for references, ethical considerations, conflicts-of-interest, as well as data, and code availability statements.
%For the final versions, authors of accepted papers will be given 1 extra content page to take the reviews into account.}
%\todo[inline]{Atenção as "Language Resource References", para já estão todas na "Bibliographical References".}

\section{Introduction}

The present paper contributes foundation models that represent the development and the populating of the still very scarce ecosystem of fully open large language models of the encoder family of Transformers specifically developed for the Portuguese language, that is models that are open source and openly distributed with for free with an open license.

Since their appearance in \citep{Vaswani:2017:Transformer} and given their superior performance vis a vis their viable alternatives, neural language models based on the Transformer architecture became the mainstream approach for virtually any natural language processing task \citep{Brown:2020:GPT3,Raffel:2020:T5,He:2021:DeBERTa}.
Transformers were proposed in an encoder-decoder setup \citep{Raffel:2020:T5}, but encoder-only and decoder-only setups have also been shown highly competitive by subsequent research \citep{Devlin:2019:BERT,He:2021:DeBERTa,Brown:2020:GPT3}. %with decoders becoming specially notable with ChatGPT \citep{Ouyang:2022:InstructGPT,openai2023gpt4}.

Despite the outstanding visibility that the Transformer-based decoder models have deservedly garnered, especially with the availability of ChatGPT for the general public, the models of the encoder family have not lost their traction as they have maintained a competitive performance in non-generative tasks, especially in those tasks primarily related to classification \citep{He:2021:DeBERTa,Zhong:2022:Vega}.\footnote{At the time of writing, as a way of confirmation of this remark, the top performing model in the SuperGLUE benchmark (\url{https://super.gluebenchmark.com/leaderboard}) is an encoder, namely the Vega v2 model \citep{Zhong:2022:Vega}.}

The largest and more powerful foundation models have been developed for English --- \citep{He:2021:DeBERTa,Touvron:2023:Llama2} among many others ---, which is the language that, among the more than 7~000 idioms on the planet, is by a very large margin the one whose research is better funded,  better technologically prepared for the digital age and for which more language resources have been developed \citep{rehm2023}. %An emblematic example of this state of affairs is the release, while we were writing the present paper, of Google's state of the art, top performing decoders Gemma, ``trained on 2T and 6T tokens respectively of primarily-English data from web documents, mathematics, and code'' \citep{google2024gemma}.

Additionally, multilingual models have also been developed, whose training is done over datasets that extend its majority of English data with proportionally much smaller data portions from a few other languages \citep{Devlin:2019:BERT,Chowdhery:2022:Palm,Scao:2022:Bloom}.
Leveraged by the sheer volume of data thus made available, these models have shown competitive performance in handling tasks in the languages, other than English, whose data portions are a minority in their training set \citep{wu-dredze-2019-beto}. 

On par with these results and their relevance for some multilingual natural language tasks, especially machine translation, other approaches have been explored, namely with the continuation of the pre-training of multilingual or plain English models with data from a specific language.
Reported results seem to converge in indicating that when their continued training is appropriately setup, the performance of the resulting models on language-specific tasks shows important improvements over a possible baseline model whose training was performed from scratch with the same (comparatively small) amount of language-specific data \citep{kim-etal-2021-changes,pires2023sabia,Rodrigues:2023:Albertina}.

Adopting this latter approach and adding to the previous work on the neural encoding of Portuguese \citep{Rodrigues:2023:Albertina,Sousa:2020:BERTimbau}, the present paper puts forward further models for this language that expand its ecosystem of open encoders.
These encoders cumulatively comply with all the features of being open source, publicly available for free, and distributed under a most permissive license (including for research and for commercial purposes).
Furthermore, they are available for two variants of Portuguese: European Portuguese, spoken in Portugal (PTPT), and American Portuguese, spoken in Brazil (PTBR).

Taking as reference the existing state-of-the-art 900 million parameter encoder Albertina \citep{Rodrigues:2023:Albertina}, which complies with all the above requirements, in this paper we present the extension of the ecosystem of open encoders for Portuguese with a larger, top performance-driven encoder model with 1.5~billion parameters, \Modellarge, and a smaller, efficiency-driven encoder model with 100~million parameters, \Modelsmall.
These models are distributed from \url{https://huggingface.co/PORTULAN}.

While achieving these central goal, further results that are relevant for this ecosystem were obtained as well: new datasets for Portuguese based on the trusted GLUE \citep{Wang:2018:GLUE} and SuperGLUE \citep{Wang:2019:SuperGLUE} benchmarks, which are distributed openly; and state-of-the-art performance for Portuguese in various natural language processing tasks in these benchmarks.

The remainder of this paper is structured as follows:
the next Section~\ref{sec:rel_work} discusses related work.
In Section~\ref{sec:data} the data used in the creation of the various models is presented;
the encoder models created in this study are described in Section~\ref{sec:models};
Section~\ref{sec:results} presents the evaluation results; and
Section~\ref{sec:conclusion} closes the paper with concluding remarks.

\section{Related work}
\label{sec:rel_work}

The advent of the Transformer architecture \citep{Vaswani:2017:Transformer} represents a revolutionary milestone in the field of Natural Language Processing.
With its attention mechanisms, the Transformer enabled the efficient modeling of contextual information in text, paving the way for the development of powerful models. 

The success of this architecture led to the emergence of various encoder models, such as BERT \citep{Devlin:2019:BERT}, RoBERTa \citep{Liu:2019:roberta}, and DeBERTa \citep{He:2021:DeBERTa}, which set new standards for language comprehension tasks. Nevertheless, they cater exclusively for the English language.

To address linguistic diversity, multilingual encoder models emerged as a promising solution.
Notable examples include mBERT \citep{Devlin:2019:BERT}, XLM \citep{Conneau:2019:XLM} and XLM-R \citep{Conneau:2020:XLM-R}, among others, which support multiple languages and seek to bridge language barriers.

In contrast, a few encoder models that cater for specific languages have also been introduced. 
For instance, 
%BERTa \citep{Armengol:2021:multilingualmodels} for Catalan, 
ERNIE \citep{Sun:2021:ernie3} for Chinese, 
%BERTje \citep{DeVries:2019:bertje} for Dutch, 
%FinBERT \citep{Virtanen:2019:bertfinnish} for Finnish, 
CamemBERT \citep{Martin:2020:Camembert} for French, 
and MarIA \citep{Gutierrez:2022:maria} for Spanish, among others. 
These have demonstrated the importance of language-tailored models in capturing language-specific nuances, which multilingual models cannot so easily ensure \citep{papadimitriou:2023:mbertaccent}.

Concerning Portuguese, previous encoder models such as the 900 million parameter Albertina \citep{Rodrigues:2023:Albertina} and the 335 million parameter BERTimbau \citep{Sousa:2020:BERTimbau} have made significant contributions.
With BERTimbau covering PTBR, and Albertina covering both PTPT and PTBR variants, these models have not only bolstered the Portuguese NLP ecosystem but have also set the path for the development of more advanced language models tailored to the Portuguese language.

In this paper, we aim at adding to this existing work by contributing further encoder models with further dimensions, also covering both the European PTPT and the American PTBR variants of Portuguese.

\section{Data}
\label{sec:data}

In this section, we present the data used for the training and testing of our encoder models.

In both their variants, PTBR and PTPT, for our smaller, 100 million parameter model, we resort to the Portuguese subset of the OSCAR dataset \citep{Abadji:2022:Oscar}. And for our larger, 1.5 billion parameter model, we resort to the Portuguese subset of the CulturaX dataset \citep{Nguyen:2023:CulturaX}.
Additionally, for the models handling the PTPT variants, the dataset we used included also the monolingual corpora DCEP, ParlamentoPT and Europarl \citep{Hajlaoui:2014:DCEP,Koehn:2005:Europarl, Rodrigues:2023:Albertina}.

These corpora and their curation are described in detail below in the next Subsection, and their sizes are summarized in Table~\ref{tab:aggregate-datasets}.

\begin{table}
    \centering
    \begin{tabular}{lrr}
        \toprule
        dataset         & exs (M)& words (B)\\ \midrule
        \Modelsmall PT  & 10.2        &  2.4   \\  % OSCAR + DCEP + ParlamentoPT + europarl
        \Modelsmall BR  &  4.1        &  2.7   \\  % OSCAR
        \Modellarge PT  & 16.1        &  4.3   \\  % CulturaX + DCEP + ParlamentoPT + europarl
        \Modellarge BR  & 87.9        & 36.2   \\  % CulturaX
        \bottomrule
    \end{tabular}
    \caption{Size of datasets used for training, in millions of examples (exs) and in billions of words.}
    \label{tab:aggregate-datasets}
\end{table}

\subsection{Training Data}

While both multilingual datasets, OSCAR and CulturaX, distribute their Portuguese subsets separately, they do not provide further separation between European Portuguese and American Portuguese within these subsets. 
To separate the texts in one variant from the texts in the other, we use the source URLs provided with every data entry and filter by top-level domain. We only keep entries with the ``.br'' top-level domain, and add them to the PTBR subset, and with the ``.pt'' top-level domain, for the PTPT subset.

From these datasets, data entries of domains whose content should not be redistributed were removed, in order to limit the possibility of content reproduction by the models or by future derivatives that will resort to these datasets. %which we distribute together with our models for the sake of scientific reproduction.\todo{indicar url, João Silva sabe indicar}\footnote{The domains filtered out include periodicals, among others, and can be consulted at <url redacted for anonymization>.}
%\todo{Why filter out news domains? Encoder cannot reproduce the content they were trained with! Redo?}

\paragraph{OSCAR Corpus}
The project promoting the OSCAR corpus is an open source project which distributes multilingual datasets for machine learning and artificial intelligence applications \cite{Abadji:2022:Oscar}.

The OSCAR subset for Portuguese we use is based on November/December 2022 version of Common Crawl, which is an automatic crawl from the web. Despite being a crawl, the final dataset is of relatively good quality due the filtering performed on the corpus by its authors.
As can be seen in Table~\ref{tab:datasets}, we end up with subsets of OSCAR for the two Portuguese variants that have a not too distinct number of examples and words.

% the contains [...] documents. After filtering by top-level domain, we end up with [...] documents for European Portuguese and [...] documents for American Portuguese (the remaining documents were discarded).
% Finally, after the quality filtering, the European dataset has [...] documents and the American Portuguese [...] documents, which corresponds to, [...] tokens for European Portuguese and [...] tokens for American Portuguese.

\paragraph{CulturaX Corpus}
CulturaX is a multilingual corpus, freely available for research and AI development \cite{Nguyen:2023:CulturaX}, created by combining and extensively cleaning two other large datasets, mC4 \citep{Xue:2021:MC4} and OSCAR.

The CulturaX subset for PTBR is an order of magnitude larger than for PTPT, as depicted in Table \ref{tab:datasets}, both in examples and words. 
This does not present itself as a problem since we aim to develop the best model possible for each variant. 
% After filtering by top-level domain, we end up with [...] documents for European Portuguese and [...] documents for American Portuguese (again, the remaining documents were discarded).
% Finally, after the quality filtering the European dataset has [...] documents and the American Portuguese [...] documents, which corresponds to, [...] tokens for European Portuguese and [...] tokens for American Portuguese.

%\todo[inline]{CulturaX BR tem 87,930,667 exemplos. CulturaX PT tem 8,885,138 exemplos.}
%\todo[inline]{Tokens usados para treinar: só estimativa, número real não dá porque cada documento foi truncado e alguns usaram padding. Estimativa usando  seq.length vezes total batches vezes steps }

\paragraph{Other Corpora}

In addition to the above language resources, for the European Portuguese versions we also include in our training set: 
(i)~the Portuguese portion of DCEP \citep{Hajlaoui:2014:DCEP}, a Digital Corpus of the European Parliament;
(ii)~the Portuguese portion of Europarl \citep{Koehn:2005:Europarl}, the European Parliament Proceedings Parallel Corpus;
and (iii)~ParlamentoPT \citep{Rodrigues:2023:Albertina}, a corpus of transcriptions of the debates in the Portuguese Parliament.

These corpora are based on human transcriptions of parliamentary debates and can be assumed to be of very high quality, despite their limited domain.
They provide a good complement to OSCAR and CulturaX.

% The DCEP corpus contains [...] documents, after the quality filtering we end up with [...] documents, which corresponds to [...] tokens.

% The Europarl corpus contains [...] documents, after the quality filtering we end up with [...] documents, which corresponds to [...] tokens.

% The ParlamentoPT corpus contains [...] documents, after the quality filtering we end up with [...] documents, which corresponds to [...] tokens.

%\todo[inline]{DCEP tem 2,502,119 examples. ParlamentoPT tem 2,891,520 exemplos. Europarl tem 1,827,984 exemplos.}

%\todo[inline]{Não é possivel saber o número de tokens certo, só estimativa tendo em conta os batches e seq. length.}

\paragraph{}
\noindent
Finally, we apply further quality filtering to all corpora---except to CulturaX, since it already has a good quality filtering step---, through the use of the Bloom pre-processing pipeline \citep{Laurenccon:2022:Bloom_filter}.

Table~\ref{tab:datasets} presents statistics for all the corpora used in this work; all these numbers are calculated right before training the model, i.e. after splitting between variants and applying all types of additional content filtering.

% OSCAR BR      4,098,364 exemplos, 2,728,209,440 palavras
% OSCAR PT      3,024,095 exemplos, 1,975,566,706 palavras
% CulturaX BR  87,930,667 exemplos, 36,200,797,106 palavras
% CulturaX PT   8,885,138 exemplos, 3,895,670,848 palavras
% DCEP          2,502,119 exemplos, 75,990,452 palavras
% ParlamentoPT  2,891,520 exemplos, 288,981,563 palavras
% Europarl      1,827,984 exemplos, 49,303,374 palavras
\begin{table}
    \centering
    \begin{tabular}{lrr}
        \toprule
        dataset       &  examples (M)& words (M)\\ \midrule
        OSCAR ptbr    &  4.1         & 2,728    \\
        OSCAR ptpt    &  3.0         & 1,976    \\ \midrule
        CulturaX ptbr & 87.9         & 36,201   \\
        CulturaX ptpt &  8.9         & 3,896    \\ \midrule
        DCEP          &  2.5         & 76       \\
        ParlamentoPT  &  2.9         & 289      \\
        Europarl      &  1.8         & 49       \\
        \bottomrule
    \end{tabular}
    \caption{Number of examples and words for each dataset for training}
    \label{tab:datasets}
\end{table}

%\paragraph{Training data statistics}
%Taking all this into account:
%(i)~the European Portuguese variant of the \Modelsmall is trained on [...] tokens;
%(ii)~the American Portuguese variant of the \Modelsmall is trained on [...] tokens;
%(iii)~the European Portuguese variant of the \Modellarge is trained on [...] tokens;
%(iv)~the American Portuguese variant of the \Modellarge is trained on [...] tokens;

\subsection{Testing Data}\label{sec:downstream-tasks}

The performance of encoder models are typically undertaken by testing them in downstream tasks.
For the Portuguese language, both variants, there is however a lack of such datasets, either in quality or in quantity, to appropriately evaluate an encoder models.
The only dataset created from scratch in (American) Portuguese, that we could find, is the ASSIN 2 dataset \citep{real2020assin} that was used to evaluate BERTimbau.    

To cope with this hindrance, we contribute new test datasets for Portuguese based on the GLUE \citep{Wang:2018:GLUE} and SuperGLUE \citep{Wang:2019:SuperGLUE} benchmarks.

We obtain these datasets through machine translation from English using DeepL,\footnote{\url{https://www.deepl.com/}} which allows translation either to PTPT or to PTBR, and is regarded as one of the best machine translation services available online.\footnote{The construction is thoroughly presented in \citep{extraGlue2023}}
%\todo{colocar o paper do extraglue no arxiv? falar com Henrique}

The exception to this translation process, concerns the PTBR portion of GLUE, which we took from PLUE \citep{Gomes:2020:plue}, to avoid redoing valid work already present in the literature and openly distributed.

\paragraph{ASSIN 2 tasks}

The ASSIN 2 dataset contains two tasks:
(i)~RTE, for recognizing textual entailment,
and
(ii)~STS, for semanting textual similarity.

\paragraph{GLUE tasks}

From GLUE we chose four tasks: 
two similarity tasks, 
(i)~MRPC, for detecting whether two sentences are paraphrases of each other, 
and 
(ii)~STS-B, for semantic textual similarity;
and two inference tasks, 
(iii)~RTE, for recognizing textual entailment, 
and 
(iv)~WNLI, for coreference and natural language inference. 

\paragraph{SuperGLUE tasks}

As for SuperGlue, we also chose four tasks: 
two QA tasks, 
(i)~MultiRC, for detecting whether an answer to a question about a paragraph is correct or not, 
and 
(ii)~BoolQ, for answering \textit{yes} or \textit{no} to a question about a passage; 
one reasoning task, 
(ii)~COPA, given a premise sentence and two possible choices, the system must determine either the cause or effect of the premise from two possible choices; 
and one inference task with three labels,
(iv)~CB, for predicting how much the text commits to the clause.

\section{Models}
\label{sec:models}
This section describes the training of the models contributed in this paper.

\subsection{The starting models}

We use DeBERTa \citep{He:2021:DeBERTa} as a starting point from which to continue the pre-training of our models over Portuguese data.
This is an encoder that incorporates a new attention mechanism, making it particularly effective for a wide range of natural language processing tasks.
DeBERTa's architecture disentangles attention patterns, improving its ability to capture relationships between words and phrases in a text.

With its different model sizes, including the compact DeBERTa-Base with 100 million parameters, the DeBERTa-XLarge with 900 million parameters, and the high-capacity DeBERTa-XXLarge with 1.5 billion parameters, it caters for various NLP requirements.

The only encoder for both variants PTP and PTBR variants of Portuguese, the existing 900 million parameter model Albertina, was obtained by continuing the pre-training of DeBERTa-XLarge with Portuguese \citep{Rodrigues:2023:Albertina}.

With the same goal in mind, we start from the DeBERTa-Base to construct our \Modelsmall \xspace models, and from the DeBERTa-XXLarge, for our \Modellarge \xspace models.

\subsection{The \Modelsmall \xspace foundation model}

The two smaller models, \Modelsmall PT and \Modelsmall BR, are constructed upon the DeBERTa Base V1 model,
%\todo{JS: Acima chamámos DeBERTa-Base}
comprising 100 million parameters. 

The models were trained on a a2-megagpu-16gb Google Cloud A2 node equipped with 16 GPUs, 96 vCPUs, and 1.360 GB of RAM, and their training took approximately one day of compute.
This configuration resulted in a batch size of 3072 samples, with 192 samples allocated per GPU, when trying to fill the whole memory available.

We used the original DeBERTa tokenizer for both models, implementing a 128-token sequence truncation and dynamic padding.
The training was performed under a learning rate of 1e-5, with linear decay and 10k warm-up steps, determined after a few exploratory trials.
The PTPT model underwent 200 training epochs, while the PTBR model underwent 150, accumulating roughly 180k training steps in each case.

\subsection{The \Modellarge \xspace foundation model}

As for the larger models, \Modellarge PT and PTBR, we developed them upon the DeBERTa XXLarge V2 encoder, comprising 1.5 billion parameters.

Similarly to the smaller models, the two \Modellarge models were trained on a a2-megagpu-16gb Google Cloud A2 node. 

We resorted to the original DeBERTa V2 tokenizer for both models, implementing a 128-token sequence truncation and dynamic padding for 250k steps, a 256-token sequence-truncation for 80k steps and finally a 512-token sequence-truncation for 60k steps. 
These steps correspond to the equivalent setup of 48 hours on a2-megagpu-16gb Google Cloud A2 node for the 128-token input sequences, 24 hours of computation for the 256-token input sequences and 24 hours of computation for the 512-token input sequences.

We applied a learning rate of 1e-5, with linear decay and 10k warm-up steps, determined after a few exploratory trials

\section{Evaluation and discussion}
\label{sec:results}

This section presents and discusses the evaluation of our models, introduced just above in Section~\ref{sec:models}, with respect to the downstream tasks, introduced in Section~\ref{sec:downstream-tasks}, after their fine-tuning on these tasks.

Additionally, for the sake of a thorough comparative evaluation of these models,  this section also presents the results of fine-tuning and evaluating in the same downstream tasks, the pre-existing models in the ecosystem of encoders for Portuguese, namely the 900 million parameter Albertina and the 335 million parameter BERTimbau We also evaluate with the two DeBERTa baseline models, with 100 million and 1.5 billion parameter, trained mostly with English data, which we did not continue the training on further Portuguese data. 

The compilation of all these results are in Table~\ref{tab:results_ptbr}, for the model versions concerning the PTBR variant, and Table~\ref{tab:results_ptpt}, for the PTPT variant.

\begin{table*}
\centering
\begin{tabular}{lcc}
\toprule
                      &\multicolumn{2}{c}{ASSIN2} \\ 
model                          &RTE            &STS \\
\cmidrule(r){1-1}\cmidrule(lr){2-3}
\Modellarge BR L      &\textbf{0.9153}&        0.8647  \\
\Modellarge BR S      &        0.9109 &\textbf{0.8688} \\
Albertina 900M PTBR   &        0.9130 &        0.8676  \\
BERTimbau (335M)      &        0.8913 &        0.8531  \\
\Modelsmall BR        &        0.8747 &        0.8269  \\
\midrule
DeBERTa 1.5B EN        &        0.8803 &        0.8356  \\
DeBERTa 100M EN        &        0.8369 &        0.7760  \\
\bottomrule
\end{tabular}
\caption{Evaluation scores for \textbf{PTBR} on the ASSIN2 native American Portuguese dataset. Performance on RTE is measured with accuracy and on STS with Pearson.}
\label{tab:results_ptbr_assin2}
\end{table*}

\begin{table*}
\centering
\begin{tabular}{lcccccccc}
\toprule
                      &\multicolumn{4}{c}{GLUE}   &\multicolumn{4}{c}{SuperGLUE}\\ 
model                          &RTE            &WNLI           &MRPC           &STS-B          &COPA           &CB             &MultiRC        &BoolQ  \\
\cmidrule(r){1-1}\cmidrule(lr){2-5}\cmidrule(l){6-9}
% valores submetidos ao LREC (ainda não tinha terminado o grid search):
%\Modellarge BR        &\textbf{0.8628}&        0.3803 &        0.8610 &\textbf{0.8993}&\textbf{0.8900}&\textbf{0.6130}&        0.3639 &\textbf{0.8498} \\
\Modellarge BR L      &\textbf{0.8676}&        0.4742 &        0.8622 &\textbf{0.9007}&        0.7767 &        0.6372 &\textbf{0.7667}&\textbf{0.8654} \\
\Modellarge BR S      &        0.8123 &        0.4225 &        0.8638 &        0.8968 &\textbf{0.8533}&\textbf{0.6884}&        0.6799 &        0.8509  \\
% valores submetidos ao LREC (ainda não tinha terminado o grid search; apenas COPA, CB, MultiRC e BoolQ foram grid-searched; os valores de RTE, WNLI, MRPC e STS-B vieram do primeiro artigo do albertina):
%Albertina PTBR (900M) &        0.7545 &        0.4601 &\textbf{0.9071}&        0.8910 &        0.7300 &        0.4739 &        0.6705 &        0.8437 \\
% valores finais:
Albertina 900M PTBR   &        0.7545 &        0.4601 &\textbf{0.9071}&        0.8910 &        0.7767 &        0.5799 &        0.6731 &        0.8385 \\
% valores submetidos ao LREC (ainda não tinha terminado o grid search):
%BERTimbau (335M)      &        0.6446 &\textbf{0.5634}&        0.8873 &        0.8842 &        0.7200 &        0.5087 &        0.6724 &        0.7734 \\
% valores finais (apenas COPA, CB, MultiRC e BoolQ foram grid-searched; os valores de RTE, WNLI, MRPC e STS-B vieram do primeiro artigo do albertina:
BERTimbau (335M)      &        0.6446 &        0.5634 &        0.8873 &        0.8842 &        0.6933 &        0.5438 &        0.6787 &        0.7783 \\
% valores submetidos ao LREC (ainda não tinha terminado o grid search):
%\Modelsmall BR        &        0.6570 &\textbf{0.5634}&        0.8468 &        0.8467 &         n.a.  &        0.4734 &        0.6517 &        0.7615 \\
% valores finais:
\Modelsmall BR        &        0.6582 &\textbf{0.5634}&        0.8149 &        0.8489 &         n.a.  &        0.4771 &        0.6469 &        0.7537 \\
\midrule
% valores submetidos ao LREC (ainda não tinha terminado o grid search):
%DeBERTa1.5B           &        0.7112 &\textbf{0.5634}&        0.8545 &        0.0123 &        0.5700 &        0.4307 &        0.3639 &        0.6217 \\
% valores finais:
DeBERTa 1.5B EN        &        0.7810 &        0.4789 &        0.8555 &        0.8600 &        0.4733 &        0.4648 &        0.6738 &        0.8315 \\
% valores submetidos ao LREC (ainda não tinha terminado o grid search):
%DeBERTa100M           &        0.5957 &\textbf{0.5634}&        0.8080 &        0.8265 &         n.a.  &        0.4739 &        0.6395 &        0.6884 \\
% valores finais:
DeBERTa 100M EN        &        0.5716 &        0.5587 &        0.8060 &        0.8266 &         n.a.  &        0.4739 &        0.6391 &        0.6838 \\
\bottomrule
\end{tabular}
\caption{Evaluation scores for \textbf{PTBR}. Performance on RTE, WNLI, BoolQ and COPA is measured with accuracy, on MRPC, MultiRC and CB with F1, and on STS-B with Pearson.}
\label{tab:results_ptbr}
\end{table*}

\begin{table*}
\centering
\begin{tabular}{lcccccccc}
\toprule
                     &\multicolumn{4}{c}{GLUE}   &\multicolumn{4}{c}{SuperGLUE}\\ 
model                 &        RTE    &        WNLI   &        MRPC   &        STS-B  &        COPA   &        CB     &       MultiRC &        BoolQ  \\ \cmidrule(r){1-1}\cmidrule(lr){2-5}\cmidrule(l){6-9}
% valores submetidos ao LREC (ainda não tinha terminado o grid search):
%\Modellarge PT        &\textbf{0.8917}&        0.5070 &        0.9204 &        0.8766 &\textbf{0.8600}&\textbf{0.6140}&        0.6723 &\textbf{0.8498}\\
\Modellarge PT L    &\textbf{0.8809}&        0.4742 &        0.8457 &\textbf{0.9034}&\textbf{0.8433}&\textbf{0.7840}&\textbf{0.7688}&\textbf{0.8602}\\
\Modellarge PT S    &        0.8809 &        0.5493 &        0.8752 &        0.8795 &        0.8400 &        0.5832 &        0.6791 &        0.8496 \\
% valores submetidos ao LREC (ainda não tinha terminado o grid search):
%Albertina PTPT(900M)  &        0.8339 &        0.4225 &\textbf{0.9171}&        0.8801 &        0.7300 &        0.4739 &        0.6782 &        0.8437 \\
% valores finais (RTE, WNLI, MRPC e STS-B copiados do primeiro artigo do albertina):
Albertina 900M PTBR   &        0.8339 &        0.4225 &\textbf{0.9171}&        0.8801 &        0.7033 &        0.6018 &        0.6728 &        0.8224 \\
% valores submetidos ao LREC (ainda não tinha terminado o grid search):
%\Modelsmall PT        &        0.5848 &\textbf{0.5634}&        0.8793 &        0.8624 &        n.a.   &        0.4734 &        0.6564 &        0.7700 \\ % valores finais:
\Modelsmall PT        &        0.6919 &        0.4742 &        0.8047 &        0.8590 &        n.a.   &        0.4529 &        0.6481 &        0.7578 \\ \midrule
% valores submetidos ao LREC (ainda não tinha terminado o grid search):
% DeBERTa1.5B           &        0.8087 &        0.4554 &\textbf{0.9220}&        0.8554 &        0.4800 &        0.4492 &        0.6538 &        0.8291 \\
% valores finais:
DeBERTa 1.5B EN         &        0.8147 &        0.4554 &        0.8696 &        0.8557 &        0.5167 &        0.4901 &        0.6687 &        0.8347 \\
% valores submetidos ao LREC (ainda não tinha terminado o grid search):
%DeBERTa 100M           &        0.6029 &\textbf{0.5634}&        0.8722 &        0.8241 &        n.a.   &        0.4612 &        0.6406 &        0.6596 \\
% valores finais:
DeBERTa 100M EN         &        0.6029 &\textbf{0.5634}&        0.7802 &        0.8320 &        n.a.   &        0.4698 &        0.6368 &        0.6829 \\
\bottomrule
\end{tabular}
\caption{Evaluation scores for \textbf{PTPT}. Performance on RTE, WNLI, BoolQ and COPA is measured with accuracy, on MRPC, MultiRC and CB with F1, and on STS-B with Pearson.}
\label{tab:results_ptpt}
\end{table*}

\subsection{Fine-tuning}
% Fine-tuning on downstream tasks is a pivotal step in maximizing the performance of pre-trained language models. To further enhance the efficacy of this process, a critical component is performing a hyper-parameter search on validation sets. This involves systematically exploring and optimizing hyper-parameters such as learning rates, batch sizes, and the number of training epochs on the validation data.

% Hyper-parameter search fine-tunes the model's parameters, ensuring it aligns optimally with the specific task's requirements. By leveraging the validation set, which serves as a representative dataset for the target task, it enables the selection of the most suitable hyper-parameter values that lead to improved model performance.

% Incorporating hyper-parameter search into the fine-tuning process allows for a more fine-grained and precise adaptation of the model to the downstream task, ultimately leading to better results and more efficient utilization of the pre-trained model's capabilities.
%\todo{AHB: refazer: Estes 3 parágrafos mastigam todos a mesma ideia, ainda por cima a um nível demasiddo "pedagógio/escolar" para o nível dos leitores do paper...}

Each model under evaluation was fine-tuned on each of the eight downstream tasks obtained from GLUE and SuperGLUE and introduced in Section~\ref{sec:downstream-tasks}.\footnote{The exception were the 100M DeBERTa models (DeBERTa-base and both versions of \Modelsmall), which were not evaluated on the COPA task because the Hugging Face head for multiple choice does not support DeBERTa v1 models.} In order to proceed with hyper-parameter optimization, the following hyper-parameter values were chosen for our grid-search:
%\vspace{0.5cm}
\begin{itemize} \setlength\itemsep{0em} % ajuste rápido para ter uma lista um pouco mais compacta
    \item Epochs: $5$
    \item Batch size: $4$
    \item Learning rate: $\{1 \times 10^{-5}, 5 \times 10^{-5}, 1 \times 10^{-6}\}$
    \item Learning rate scheduler type: linear
    \item Warm up ratio: $0.1$  
    \item Adam epsilon: $1 \times 10^{-6}$
    \item Weight decay: $0.01$
    \item Dropout: $\{0, 0.1\}$
    \item BF16: $\{0, 1\}$
\end{itemize}
%As for non changing hyper-parameters we chose a batch size of 4 (with gradient accumulation of 2 and per device batch size of 2), and we train all models for 5 epochs, 

% 4104 = 12x(5x10+2x9+4x8+2x7)x3
%        3 random seeds
%        12 hp combinations
%        5 PT-BR >100M models fine tuned in 10 PT-BR tasks
%        2 PT-BR 100M models fine tuned in 9 PT-BR tasks (all except COPA)
%        4 PT-PT >100M models fine tuned in 8 PT-PT tasks
%        2 PT-PT 100M models fine tuned in 7 PT-PT tasks (all except COPA)

A hyper-parameter grid search was performed for each pre-trained model/task combination, resulting in a total of \textbf{4104} fine-tuned and evaluated models. This number results from 12 combinations of hyper-parameter values (3 learning rates $\times$ 2 dropout values $\times$ 2 BF16 values), times the number of tasks (10 for PT-BR and 8 for PT-PT), times the number of evaluated pre-trained models\footnote{The 100M parameter models could not be evaluated in the COPA task for lack of support for these models in the HuggingFace head implementation for this task.} (7 for PT-BR and 6 for PT-PT), times 3 random seeds.

As presented in Section~\ref{sec:data}, the GLUE and SuperGLUE evaluation datasets were translated into both Portuguese variants from their English originals.

It is noteworthy that the test sets from from the GLUE and SuperGLUE datasets are not distributed with ground labels, as evaluation is setup to proceed by submitting online the data to be evaluated.
Given that the number of such online submissions per month for each user is highly limited and very small, and given the very large number of models and tasks and thus of evaluation runs we needed to cope with, it was not practically viable to resort to such online evaluation service.
As a consequence, to proceed with our very large experimental space, we adopted the same methodology as we did for the 900 million parameter Albertina \cite{Rodrigues:2023:Albertina}: we used the validation partitions of the downstream datasets for testing; and for training, we randomly split the partition that is originally distributed for training into 90\% that we used for actual training and into the remaining 10\% that we used for development and validation purposes.

%We test all models on the validation portion of the datasets, since test set labels are not available for GLUE and SuperGLUE.\footnote{We do not submit the GLUE and SuperGLUE benchmarks because we do not have all the tasks translated to Portuguese, and due to the restrictions on the number of submissions allowed to each benchmark.}
%As for training, we randomly split the training set into 90\% for training and the remaining 10\% for development purposes.

After acquiring the best hyper-parameter values on the data that were set aside for development purposes and by using such hyper-parameters, the performance scores were obtained by testing on the subsets that were left for evaluation, which are displayed in
Tables~\ref{tab:results_ptbr_assin2}, ~\ref{tab:results_ptbr} and~\ref{tab:results_ptpt}. The values presented are the average scores of 3 runs with different random seeds.

\subsection{\Modellarge \xspace fine-tuned}
%\todo{rever este paragráfo que acrescentei (LMG) a explicar o significado das duas linhas para o modelo 1.5B em cada tabela}
Since most tasks have input sizes closer to 256 than to 512, we evaluated two variants of the \Modellarge \xspace model: the models with suffix S (short) in Tables~\ref{tab:results_ptbr} and~\ref{tab:results_ptpt} are fine-tuned from checkpoints after pre-training with sequences of 256 tokens; while the models with suffix L (long) are fine-tuned from the final checkpoints, i.e. after pre-training with sequences of 512 tokens.

In almost all tasks and for both language variants, our largest model, with 1.5 billion parameters, shows the best performance scores, and in the few cases where that is not the case, it competitively come close to the best scoring model.

It is of note that among the downstream tasks, WNLI appears somehow as an outlier as the performance level of the different models on it is not aligned with their performance level in the other tasks. This has been already observed also with Albertina 900 M \cite{Rodrigues:2023:Albertina}, which attributed this to the very small size of the WNLI dataset.

In its overall performance, this largest model surpasses the previously best model Albertina 900M in this ecosystem, and offers thus the state-of-the-art performance in most tasks for Portuguese by an open encoder.

\subsection{ \Modelsmall \xspace fine-tuned}

With 100 million parameters, our \Modelsmall \xspace model is the smallest in this ecosystem of open encoders for Portuguese.
Yet, it has very good performance taking into account its reduced size.

Taking WNLI aside, \Modelsmall \xspace matches or surpasses its base model (DeBERTa 100M) in all 16 tasks, except in CB for PTPT.

On the other hand, our \Modelsmall BR is very competitive with respect to the BERTimbau model, whose 335 million parameters are more than the triple of its size. It surpasses BERTimbau's performance in GLUE's RTE, and supports a very competitive second position in most of the other tasks. Likely, this is the consequence of BERTimbau having BERT \cite{Devlin:2019:BERT} as its base model, while \Modelsmall is based in the more advanced DeBERTa \cite{He:2021:DeBERTa}.

\subsection{Discussion}

\paragraph{The larger the better}
Taking a broad view of the results in Tables \ref{tab:results_ptbr} and \ref{tab:results_ptpt}, overall and as expected, the larger the Albertina model the better is its performance in downstream tasks.

In this respect, and taking aside WNLI, already commented on above, the exception to this trend is MRPC.  In this task, the 1.5B Albertina models are outperformed by the smaller 900M Albertinas.  Although we don't have a compelling explanation for this, it appears that the 900M parameter network may provide the optimal expressive power for learning this particular task and dataset, across the various model sizes under evaluation.%\todo{Reescrevi (LMG) este parágrafo com base nos novos resultados. @AHB: validar|ajustar sff}

%As seen, our new models, especially the \Modellarge models, outperform their smaller counterparts and previously existing models like Albertina and BERTimbau in most tasks, often by a significant margin, indicating the importance of model size for improving performance in NLP tasks.

\paragraph{The more monolingual the better} When compared to their respective DeBERTa baseline counterparts, our newly contributed models, \Modellarge and \Modelsmall, present superior performance in general. 

This adds to the empirical evidence in the literature, commented in Section \ref{sec:rel_work}, for the importance of continuing the pre-training of models with monolingual data for the language of interest, even if they started multilingual or were initially developed for another language. If appropriately prepared, the resulting models typically represent a better solution for that language.

Concerning the largest model Abertina 1.5B, and taking aside the WNLI outlier, it always improves over its baseline model. 

As for our smaller model Albertina 100M, the exception to this trend appears once again in WNLI, for PTPT,  and CB, by a small margin, also for PTPT.%\todo{Reescrevi (LMG) estes dois últimos parágrafos com base nos novos resultados. @AHB: validar|ajustar sff}

\paragraph{The more advanced the base model the better}Comparing the new \Modelsmall \xspace and \Modellarge \xspace models to the previously existing models, it is clear that the larger models offer improvements over smaller models as noted above. 

However, it is important also to note that the difference between the performance scores of \Modelsmall BR and of the 335M BERTimbau is rather small, which seems to suggest that the improvements in DeBERTa, on which our \Modelsmall is based, over BERT, which used as a base model by BERTimbau, have allowed for more efficient parameter utilization and improved performance in general.

\paragraph{The more language variants the better} For the same task and the same model dimension, the models for the European PTPT and American PTBR variants of Portuguese show different performance scores. While in general not representing a wide gap, these differences exist, as expected.

These differences should be attributed, for instance, to the possible different quality of the translations produced for the English datasets, depending on the Portuguese variant, and also attributed in some cases to the different sizes of the training corpora, etc. For instance, the training of the 1.5 billion model for PTBR was based on a 36.2 billion token dataset, while the same size model for PTPT resorted to a much smaller, 4.3 billion token corpus, as indicated in Table \ref{tab:aggregate-datasets}. 

From the three models with two versions, i.e. one version per variant, namely, the Albertina 100M, 900M and 1.5B models, it is the 900M one than may permit a more insightful comparison among its two variants given the conditions of their training were closer to each other, with a 2.7M and a 2.2M token training dataset for PTBR and PTPT, respectively \cite{Rodrigues:2023:Albertina}. 

Thus looking to the experimental results we obtained for the two  Albertina 900M versions, PTBR and PTPT, across the Tables \ref{tab:results_ptbr} and \ref{tab:results_ptpt}, one finds deltas, for instance, of 0.079 (accuracy) in RTE, 0.073 (F1) in COPA, or 0.022 (accuracy) in CB. This is in line with the same lessons drawn in \cite{Rodrigues:2023:Albertina}, and it is confirming its results. It is thus relevant to keep the two variants of Portuguese addressed by different model versions if possible.%\todo{ajustei (LMG) este paragrafo com base nos novos valores; @AHB: validar|ajustar sff}

\section{Conclusions}
\label{sec:conclusion}

The results reported in the present paper demonstrate that the models hereby contributed represent valuable advances for the ecosystem of fully open large language models of Portuguese.

With its 1.5 billion parameters, \Modellarge \xspace becomes the largest open encoder specifically developed for this language, and the one that better support state of the art performance in downstream tasks. 

With its 100 million parameter, \Modelsmall \xspace becomes, in turn, the smallest, appropriately curated and documented, open encoder of this ecosystem, and thus the one that ensures an encoding solution for this language that favours efficiency and is available to run in limited hardware. 

 It is also worth noting that the advancements contributed in this paper for both American and European variants of Portuguese cater for the linguistic diversity in this language, ensuring their relevance and applicability to a broad user base.

In conclusion, this paper presents a significant contribution to the field of language technology for Portuguese by introducing state-of-the-art large language models that serve the technological preparation of this language. The models are not only technically robust but also fully open, in the sense that are open source, openly distributed for free under an open license for both research and commercial purposes. They are adaptable for various applications, thus facilitating innovation and progress in the field. 

These models can be obtained from \url{https://huggingface.co/PORTULAN}.

Future work will include further expanding and updating this ecosystem of fully open encoders for Portuguese with other model dimensions, other language variants and other design features.

%As future work, other variants of the Portuguese language should be taken into account. 
%Portuguese variants from the African continent have millions of speakers, even surpassing those of European Portuguese; however, there is little data available to train a large language model in the same fashion as those described in this document.
%To further advance the field of natural language processing of Portuguese for these other variants, innovative approaches should be explored. This may involve the creation of specialized datasets, collaborative efforts with linguistic experts and native speakers, and the development of tailored pre-training techniques. By tackling this data scarcity issue head-on, we can pave the way for the development of effective and impactful large language models for these technologically underdeveloped Portuguese variants.

%\todo[inline]{JS: E as outras variantes de PT? O PT de Angola tem mais falantes, em número absoluto, do que os falantes de PT de Portugal (mas não deve ter dados para treinar modelos). No entanto, se nos posicionamos como um paladino da língua portuguesa que fomenta um ecossistema, é feio deixar de lado milhões de falantes dessa língua sem sequer lhes fazer menção.}

\section*{Acknowledgements}
This research was partially supported by:
PORTULAN CLARIN — Research Infrastructure for the Science and Technology of Language, funded by Lisboa 2020, Alentejo 2020 and FCT (PINFRA/22117/2016); ACCELERAT.AI - Multilingual Intelligent Contact Centers, funded by IAPMEI (C625734525-00462629);
ALBERTINA - Foundation Encoder Model for Portuguese and AI, funded by FCT (CPCA-IAC/AV/478394/2022);
and LIACC - Artificial Intelligence and Computer Science Laboratory (FCT/UID/CEC/0027/2020).

\citetlanguageresource{*}

\section{Bibliographical References}

\bibliographystyle{lrec-coling2024-natbib}
\bibliography{bibliographic_resources}

\begin{thebibliography}{12}
\expandafter\ifx\csname natexlab\endcsname\relax\def\natexlab#1{#1}\fi

\bibitem[{{F{\'a}bio Souza and Rodrigo Nogueira and Roberto Lotufo}(2020)}]{bertimbau}
{F{\'a}bio Souza and Rodrigo Nogueira and Roberto Lotufo}. 2020.
\newblock \href {https://huggingface.co/neuralmind/bert-large-portuguese-cased} {\emph{{BERTimbau Large}}}.
\newblock Hugging Face.

\bibitem[{Gomes(2020)}]{plue}
J. R. S. Gomes. 2020.
\newblock \href {https://huggingface.co/datasets/dlb/plue} {\emph{{PLUE}: Portuguese Language Understanding Evaluation}}.
\newblock Hugging Face.

\bibitem[{{Hajlaoui Najeh, Kolovratnik David, Vaeyrynen Jaakko, Steinberger Ralf, and Varga Dániel}(2012)}]{dcep}
{Hajlaoui Najeh, Kolovratnik David, Vaeyrynen Jaakko, Steinberger Ralf, and Varga Dániel}. 2012.
\newblock \emph{{DCEP: Digital Corpus of the European Parliament}}.
\newblock European Parliament - DG TRAD. European Parliament - DG TRAD, ISLRN \href{https://www.islrn.org/resources/823-807-024-162-2}{823-807-024-162-2}.

\bibitem[{{João Rodrigues and Luís Gomes and João Silva and António Branco and Rodrigo Santos and Henrique Lopes Cardoso and Tomás Osório}(2023{\natexlab{a}})}]{albertina-ptbr}
{João Rodrigues and Luís Gomes and João Silva and António Branco and Rodrigo Santos and Henrique Lopes Cardoso and Tomás Osório}. 2023{\natexlab{a}}.
\newblock \emph{{Albertina PT-BR}}.
\newblock PORTULAN CLARIN. distributed via PORTULAN CLARIN.
\newblock PID \href{https://hdl.handle.net/21.11129/0000-000F-FF43-7}{https://hdl.handle.net/21.11129/0000-000F-FF43-7}.

\bibitem[{{João Rodrigues and Luís Gomes and João Silva and António Branco and Rodrigo Santos and Henrique Lopes Cardoso and Tomás Osório}(2023{\natexlab{b}})}]{albertina-ptpt}
{João Rodrigues and Luís Gomes and João Silva and António Branco and Rodrigo Santos and Henrique Lopes Cardoso and Tomás Osório}. 2023{\natexlab{b}}.
\newblock \emph{{Albertina PT-PT}}.
\newblock PORTULAN CLARIN. distributed via PORTULAN CLARIN.
\newblock PID \href{https://hdl.handle.net/21.11129/0000-000F-FF42-8}{https://hdl.handle.net/21.11129/0000-000F-FF42-8}.

\bibitem[{{Julien Abadji and Pedro Ortiz Suarez and Laurent Romary and Benoît Sagot}(2023)}]{oscar}
{Julien Abadji and Pedro Ortiz Suarez and Laurent Romary and Benoît Sagot}. 2023.
\newblock \href {https://oscar-project.org/} {\emph{{OSCAR 23.01 -- Open Source Project on Multilingual Resources for Machine Learning}}}.
\newblock the OSCAR project.

\bibitem[{{Pengcheng He and Xiaodong Liu and Jianfeng Gao and Weizhu Chen}(2023)}]{deberta}
{Pengcheng He and Xiaodong Liu and Jianfeng Gao and Weizhu Chen}. 2023.
\newblock \href {https://github.com/microsoft/DeBERTa} {\emph{{DeBERTa: Decoding-enhanced BERT with Disentangled Attention}}}.
\newblock Microsoft.

\bibitem[{{Philipp Koehn}(2012)}]{europarl}
{Philipp Koehn}. 2012.
\newblock \href {https://www.statmt.org/europarl/} {\emph{{European Parliament Proceedings Parallel Corpus (v7)}}}.
\newblock EuroMatrixPlus project.

\bibitem[{Real et~al.(2020)Real, Fonseca, and Gon{\c{c}}alo~Oliveira}]{assin2}
Real, Livy and Fonseca, Erick and Gon{\c{c}}alo Oliveira, Hugo. 2020.
\newblock \href {https://huggingface.co/datasets/assin2} {\emph{ASSIN 2 (The {ASSIN 2} Shared Task: A Quick Overview)}}.
\newblock Hugging Face.

\bibitem[{{Thuat Nguyen and Chien Van Nguyen and Viet Dac Lai and Hieu Man and Nghia Trung Ngo and Franck Dernoncourt and Ryan A. Rossi and Thien Huu Nguyen}(2023)}]{culturax}
{Thuat Nguyen and Chien Van Nguyen and Viet Dac Lai and Hieu Man and Nghia Trung Ngo and Franck Dernoncourt and Ryan A. Rossi and Thien Huu Nguyen}. 2023.
\newblock \href {https://huggingface.co/datasets/uonlp/CulturaX} {\emph{{CulturaX: A Cleaned, Enormous, and Multilingual Dataset for Large Language Models in 167 Languages}}}.
\newblock Hugging Face.

\bibitem[{Wang et~al.(2019)Wang, Pruksachatkun, Nangia, Singh, Michael, Hill, Levy, and Bowman}]{superglue}
Wang, Alex and Pruksachatkun, Yada and Nangia, Nikita and Singh, Amanpreet and Michael, Julian and Hill, Felix and Levy, Omer and Bowman, Samuel. 2019.
\newblock \href {https://huggingface.co/datasets/super_glue} {\emph{Superglue: A stickier benchmark for general-purpose language understanding systems}}.
\newblock Hugging Face.

\bibitem[{Wang et~al.(2018)Wang, Singh, Michael, Hill, Levy, and Bowman}]{glue}
Wang, Alex and Singh, Amanpreet and Michael, Julian and Hill, Felix and Levy, Omer and Bowman, Samuel. 2018.
\newblock \href {https://huggingface.co/datasets/glue} {\emph{{GLUE}: A Multi-Task Benchmark and Analysis Platform for Natural Language Understanding}}.
\newblock Hugging Face.

\end{thebibliography}


\begin{thebibliography}{33}
\expandafter\ifx\csname natexlab\endcsname\relax\def\natexlab#1{#1}\fi

\bibitem[{Abadji et~al.(2022)Abadji, Ortiz~Suarez, Romary, and Sagot}]{Abadji:2022:Oscar}
Julien Abadji, Pedro Ortiz~Suarez, Laurent Romary, and Beno{\^\i}t Sagot. 2022.
\newblock \href {https://aclanthology.org/2022.lrec-1.463} {Towards a cleaner document-oriented multilingual crawled corpus}.
\newblock In \emph{Proceedings of the Thirteenth Language Resources and Evaluation Conference (LREC)}, pages 4344--4355.

\bibitem[{Brown et~al.(2020)Brown, Mann, Ryder, Subbiah, Kaplan, Dhariwal, Neelakantan, Shyam, Sastry, Askell et~al.}]{Brown:2020:GPT3}
Tom Brown, Benjamin Mann, Nick Ryder, Melanie Subbiah, Jared~D Kaplan, Prafulla Dhariwal, Arvind Neelakantan, Pranav Shyam, Girish Sastry, Amanda Askell, et~al. 2020.
\newblock Language models are few-shot learners.
\newblock \emph{Advances in Neural Information Processing Systems}, 33:1877--1901.

\bibitem[{Chowdhery et~al.(2022)Chowdhery, Narang, Devlin, Bosma, Mishra, Roberts, Barham, Chung, Sutton, Gehrmann et~al.}]{Chowdhery:2022:Palm}
Aakanksha Chowdhery, Sharan Narang, Jacob Devlin, Maarten Bosma, Gaurav Mishra, Adam Roberts, Paul Barham, Hyung~Won Chung, Charles Sutton, Sebastian Gehrmann, et~al. 2022.
\newblock Palm: Scaling language modeling with pathways.
\newblock \emph{arXiv preprint arXiv:2204.02311}.

\bibitem[{Conneau et~al.(2020)Conneau, Khandelwal, Goyal, Chaudhary, Wenzek, Guzm{\'a}n, Grave, Ott, Zettlemoyer, and Stoyanov}]{Conneau:2020:XLM-R}
Alexis Conneau, Kartikay Khandelwal, Naman Goyal, Vishrav Chaudhary, Guillaume Wenzek, Francisco Guzm{\'a}n, Edouard Grave, Myle Ott, Luke Zettlemoyer, and Veselin Stoyanov. 2020.
\newblock \href {https://aclanthology.org/2020.acl-main.747} {Unsupervised cross-lingual representation learning at scale}.
\newblock In \emph{Proceedings of the 58th Annual Meeting of the Association for Computational Linguistics}, pages 8440--8451.

\bibitem[{Conneau and Lample(2019)}]{Conneau:2019:XLM}
Alexis Conneau and Guillaume Lample. 2019.
\newblock Cross-lingual language model pretraining.
\newblock In \emph{Proceedings of the 33rd International Conference on Neural Information Processing Systems}.

\bibitem[{Devlin et~al.(2019)Devlin, Chang, Lee, and Toutanova}]{Devlin:2019:BERT}
Jacob Devlin, Ming-Wei Chang, Kenton Lee, and Kristina Toutanova. 2019.
\newblock {BERT}: Pre-training of deep bidirectional transformers for language understanding.
\newblock In \emph{Proceedings of the 2019 Conference of the North {A}merican Chapter of the Association for Computational Linguistics: Human Language Technologies, Volume 1 (Long and Short Papers)}, pages 4171--4186.

\bibitem[{Gomes(2020)}]{Gomes:2020:plue}
J.~R.~S. Gomes. 2020.
\newblock Plue: Portuguese language understanding evaluation.
\newblock \url{https://github.com/ju-resplande/PLUE}.

\bibitem[{Gutiérrez-Fandiño et~al.(2022)Gutiérrez-Fandiño, Armengol-Estapé, Pàmies, Llop-Palao, Silveira-Ocampo, Carrino, Armentano-Oller, Rodriguez-Penagos, Gonzalez-Agirre, and Villegas}]{Gutierrez:2022:maria}
Asier Gutiérrez-Fandiño, Jordi Armengol-Estapé, Marc Pàmies, Joan Llop-Palao, Joaquin Silveira-Ocampo, Casimiro~Pio Carrino, Carme Armentano-Oller, Carlos Rodriguez-Penagos, Aitor Gonzalez-Agirre, and Marta Villegas. 2022.
\newblock {MarIA}: {Spanish} language models.
\newblock \emph{Procesamiento del Lenguaje Natural}, pages 39--60.

\bibitem[{Hajlaoui et~al.(2014)Hajlaoui, Kolovratnik, V{\"a}yrynen, Steinberger, and Varga}]{Hajlaoui:2014:DCEP}
Najeh Hajlaoui, David Kolovratnik, Jaakko V{\"a}yrynen, Ralf Steinberger, and Daniel Varga. 2014.
\newblock {DCEP}---{D}igital corpus of the {E}uropean {P}arliament.
\newblock In \emph{Proceedings of the Ninth International Conference on Language Resources and Evaluation ({LREC}'14)}.

\bibitem[{He et~al.(2021)He, Liu, Gao, and Chen}]{He:2021:DeBERTa}
Pengcheng He, Xiaodong Liu, Jianfeng Gao, and Weizhu Chen. 2021.
\newblock \href {https://openreview.net/forum?id=XPZIaotutsD} {{DeBERTa}: Decoding-enhanced {BERT} with disentangled attention}.
\newblock In \emph{International Conference on Learning Representations}.

\bibitem[{Kim et~al.(2021)Kim, Kim, Lee, Lee, Kwak, Dong~Hyeon, Park, Kim, Kim, Seo, Lee, Jeong, Lee, Kim, Ko, Kim, Park, Kim, Kang, Ryu, Yoo, Chang, Suh, In, Park, Kim, Kim, Jeong, Yeo, Ham, Park, Lee, Kang, Kang, Ha, Park, and Sung}]{kim-etal-2021-changes}
Boseop Kim, HyoungSeok Kim, Sang-Woo Lee, Gichang Lee, Donghyun Kwak, Jeon Dong~Hyeon, Sunghyun Park, Sungju Kim, Seonhoon Kim, Dongpil Seo, Heungsub Lee, Minyoung Jeong, Sungjae Lee, Minsub Kim, Suk~Hyun Ko, Seokhun Kim, Taeyong Park, Jinuk Kim, Soyoung Kang, Na-Hyeon Ryu, Kang~Min Yoo, Minsuk Chang, Soobin Suh, Sookyo In, Jinseong Park, Kyungduk Kim, Hiun Kim, Jisu Jeong, Yong~Goo Yeo, Donghoon Ham, Dongju Park, Min~Young Lee, Jaewook Kang, Inho Kang, Jung-Woo Ha, Woomyoung Park, and Nako Sung. 2021.
\newblock \href {https://doi.org/10.18653/v1/2021.emnlp-main.274} {What changes can large-scale language models bring? {I}ntensive study on {H}yper{CLOVA}: Billions-scale {K}orean generative pretrained transformers}.
\newblock In \emph{Proceedings of the 2021 Conference on Empirical Methods in Natural Language Processing}, pages 3405--3424, Online and Punta Cana, Dominican Republic. Association for Computational Linguistics.

\bibitem[{Koehn(2005)}]{Koehn:2005:Europarl}
Philipp Koehn. 2005.
\newblock Europarl: A parallel corpus for statistical machine translation.
\newblock In \emph{Proceedings of Machine Translation Summit X: papers}, pages 79--86.

\bibitem[{Lauren{\c{c}}on et~al.(2022)Lauren{\c{c}}on, Saulnier, Wang, Akiki, Villanova~del Moral, Le~Scao, Von~Werra, Mou, Gonz{\'a}lez~Ponferrada, Nguyen et~al.}]{Laurenccon:2022:Bloom_filter}
Hugo Lauren{\c{c}}on, Lucile Saulnier, Thomas Wang, Christopher Akiki, Albert Villanova~del Moral, Teven Le~Scao, Leandro Von~Werra, Chenghao Mou, Eduardo Gonz{\'a}lez~Ponferrada, Huu Nguyen, et~al. 2022.
\newblock The {BigScience} {ROOTS} corpus: A 1.6 {TB} composite multilingual dataset.
\newblock \emph{Advances in Neural Information Processing Systems}, 35:31809--31826.

\bibitem[{Liu et~al.(2019)Liu, Ott, Goyal, Du, Joshi, Chen, Levy, Lewis, Zettlemoyer, and Stoyanov}]{Liu:2019:roberta}
Yinhan Liu, Myle Ott, Naman Goyal, Jingfei Du, Mandar Joshi, Danqi Chen, Omer Levy, Mike Lewis, Luke Zettlemoyer, and Veselin Stoyanov. 2019.
\newblock {RoBERTa}: A robustly optimized {BERT} pretraining approach.
\newblock \emph{arXiv preprint arXiv:1907.11692}.

\bibitem[{Martin et~al.(2020)Martin, Muller, Ortiz~Su{\'a}rez, Dupont, Romary, de~la Clergerie, Seddah, and Sagot}]{Martin:2020:Camembert}
Louis Martin, Benjamin Muller, Pedro~Javier Ortiz~Su{\'a}rez, Yoann Dupont, Laurent Romary, {\'E}ric de~la Clergerie, Djam{\'e} Seddah, and Beno{\^\i}t Sagot. 2020.
\newblock \href {https://aclanthology.org/2020.acl-main.645} {{C}amem{BERT}: a tasty {F}rench language model}.
\newblock In \emph{Proceedings of the 58th Annual Meeting of the Association for Computational Linguistics}, pages 7203--7219.

\bibitem[{Nguyen et~al.(2023)Nguyen, Van~Nguyen, Lai, Man, Ngo, Dernoncourt, Rossi, and Nguyen}]{Nguyen:2023:CulturaX}
Thuat Nguyen, Chien Van~Nguyen, Viet~Dac Lai, Hieu Man, Nghia~Trung Ngo, Franck Dernoncourt, Ryan~A Rossi, and Thien~Huu Nguyen. 2023.
\newblock {CulturaX}: A cleaned, enormous, and multilingual dataset for large language models in 167 languages.
\newblock \emph{arXiv preprint arXiv:2309.09400}.

\bibitem[{Osório et~al.(submited)Osório, Leite, Cardoso, Gomes, Rodrigues, Santos, and Branco}]{extraGlue2023}
Tomás~Freitas Osório, Bernardo Leite, Henrique~Lopes Cardoso, Luís Gomes, João Rodrigues, Rodrigo Santos, and António Branco. submited.
\newblock Extraglue datasets and models: Kick-starting a benchmark for the neural processing of portuguese.

\bibitem[{Papadimitriou et~al.(2023)Papadimitriou, Lopez, and Jurafsky}]{papadimitriou:2023:mbertaccent}
Isabel Papadimitriou, Kezia Lopez, and Dan Jurafsky. 2023.
\newblock \href {https://doi.org/10.18653/v1/2023.sigtyp-1.16} {Multilingual {BERT} has an accent: Evaluating {E}nglish influences on fluency in multilingual models}.
\newblock In \emph{Proceedings of the 5th Workshop on Research in Computational Linguistic Typology and Multilingual NLP}, pages 143--146. Association for Computational Linguistics.

\bibitem[{Pires et~al.(2023)Pires, Abonizio, Almeida, and Nogueira}]{pires2023sabia}
Ramon Pires, Hugo Abonizio, Thales~Sales Almeida, and Rodrigo Nogueira. 2023.
\newblock \href {http://arxiv.org/abs/2304.07880} {Sabi\'a: {Portuguese} large language models}.
\newblock \emph{arXiv preprint arXiv:2304.07880}.

\bibitem[{Raffel et~al.(2020)Raffel, Shazeer, Roberts, Lee, Narang, Matena, Zhou, Li, and Liu}]{Raffel:2020:T5}
Colin Raffel, Noam Shazeer, Adam Roberts, Katherine Lee, Sharan Narang, Michael Matena, Yanqi Zhou, Wei Li, and Peter~J Liu. 2020.
\newblock Exploring the limits of transfer learning with a unified text-to-text transformer.
\newblock \emph{The Journal of Machine Learning Research}, 21:5485--5551.

\bibitem[{Real et~al.(2020)Real, Fonseca, and Gon{\c{c}}alo~Oliveira}]{real2020assin}
Livy Real, Erick Fonseca, and Hugo Gon{\c{c}}alo~Oliveira. 2020.
\newblock The assin 2 shared task: a quick overview.
\newblock In \emph{Computational Processing of the Portuguese Language: 14th International Conference, PROPOR 2020, Evora, Portugal, March 2--4, 2020, Proceedings 14}, pages 406--412. Springer.

\bibitem[{Rehm and Way(2023)}]{rehm2023}
Georg Rehm and Andy Way, editors. 2023.
\newblock \href {https://doi.org/https://doi.org/10.1007/978-3-031-28819-7} {\emph{{European Language Equality: A Strategic Agenda for Digital Language Equality}}}.
\newblock Cognitive Technologies. Springer.

\bibitem[{Rodrigues et~al.(2023)Rodrigues, Gomes, Silva, Branco, Santos, Cardoso, and Os{\'o}rio}]{Rodrigues:2023:Albertina}
Jo{\~a}o Rodrigues, Lu{\'\i}s Gomes, Jo{\~a}o Silva, Ant{\'o}nio Branco, Rodrigo Santos, Henrique~Lopes Cardoso, and Tom{\'a}s Os{\'o}rio. 2023.
\newblock \href {https://arxiv.org/pdf/2305.06721.pdf} {Advancing neural encoding of {P}ortuguese with transformer {A}lbertina {PT}-*}.
\newblock In \emph{Progress in Artificial Intelligence (EPIA 2023)}.

\bibitem[{Scao et~al.(2022)Scao, Fan, Akiki, Pavlick, Ili{\'c}, Hesslow, Castagn{\'e}, Luccioni, Yvon, Gall{\'e} et~al.}]{Scao:2022:Bloom}
Teven~Le Scao, Angela Fan, Christopher Akiki, Ellie Pavlick, Suzana Ili{\'c}, Daniel Hesslow, Roman Castagn{\'e}, Alexandra~Sasha Luccioni, Fran{\c{c}}ois Yvon, Matthias Gall{\'e}, et~al. 2022.
\newblock Bloom: A 176b-parameter open-access multilingual language model.
\newblock \emph{arXiv preprint arXiv:2211.05100}.

\bibitem[{Souza et~al.(2020)Souza, Nogueira, and Lotufo}]{Sousa:2020:BERTimbau}
F{\'a}bio Souza, Rodrigo Nogueira, and Roberto Lotufo. 2020.
\newblock {BERTimbau}: Pretrained {BERT Models} for {Brazilian} {Portuguese}.
\newblock In \emph{Intelligent Systems}, pages 403--417.

\bibitem[{Sun et~al.(2021)Sun, Wang, Feng, Ding, Pang, Shang, Liu, Chen, Zhao, Lu et~al.}]{Sun:2021:ernie3}
Yu~Sun, Shuohuan Wang, Shikun Feng, Siyu Ding, Chao Pang, Junyuan Shang, Jiaxiang Liu, Xuyi Chen, Yanbin Zhao, Yuxiang Lu, et~al. 2021.
\newblock Ernie 3.0: Large-scale knowledge enhanced pre-training for language understanding and generation.
\newblock \emph{arXiv preprint arXiv:2107.02137}.

\bibitem[{Touvron et~al.(2023)Touvron, Martin, Stone, Albert, Almahairi, Babaei, Bashlykov, Batra, Bhargava, Bhosale et~al.}]{Touvron:2023:Llama2}
Hugo Touvron, Louis Martin, Kevin Stone, Peter Albert, Amjad Almahairi, Yasmine Babaei, Nikolay Bashlykov, Soumya Batra, Prajjwal Bhargava, Shruti Bhosale, et~al. 2023.
\newblock Llama 2: Open foundation and fine-tuned chat models.
\newblock \emph{arXiv preprint arXiv:2307.09288}.

\bibitem[{Vaswani et~al.(2017)Vaswani, Shazeer, Parmar, Uszkoreit, Jones, Gomez, Kaiser, and Polosukhin}]{Vaswani:2017:Transformer}
Ashish Vaswani, Noam Shazeer, Niki Parmar, Jakob Uszkoreit, Llion Jones, Aidan~N Gomez, {\L}ukasz Kaiser, and Illia Polosukhin. 2017.
\newblock Attention is all you need.
\newblock \emph{Advances in neural information processing systems}, 30.

\bibitem[{Wang et~al.(2019)Wang, Pruksachatkun, Nangia, Singh, Michael, Hill, Levy, and Bowman}]{Wang:2019:SuperGLUE}
Alex Wang, Yada Pruksachatkun, Nikita Nangia, Amanpreet Singh, Julian Michael, Felix Hill, Omer Levy, and Samuel Bowman. 2019.
\newblock Superglue: A stickier benchmark for general-purpose language understanding systems.
\newblock \emph{Advances in neural information processing systems}, 32.

\bibitem[{Wang et~al.(2018)Wang, Singh, Michael, Hill, Levy, and Bowman}]{Wang:2018:GLUE}
Alex Wang, Amanpreet Singh, Julian Michael, Felix Hill, Omer Levy, and Samuel Bowman. 2018.
\newblock {GLUE}: A multi-task benchmark and analysis platform for natural language understanding.
\newblock In \emph{Proceedings of the 2018 EMNLP Workshop BlackboxNLP: Analyzing and Interpreting Neural Networks for NLP}, pages 353--355.

\bibitem[{Wu and Dredze(2019)}]{wu-dredze-2019-beto}
Shijie Wu and Mark Dredze. 2019.
\newblock \href {https://doi.org/10.18653/v1/D19-1077} {{Beto}, {Bentz}, {Becas}: The surprising cross-lingual effectiveness of {BERT}}.
\newblock In \emph{Proceedings of the 2019 Conference on Empirical Methods in Natural Language Processing and the 9th International Joint Conference on Natural Language Processing (EMNLP-IJCNLP)}, pages 833--844, Hong Kong, China. Association for Computational Linguistics.

\bibitem[{Xue et~al.(2021)Xue, Constant, Roberts, Kale, Al-Rfou, Siddhant, Barua, and Raffel}]{Xue:2021:MC4}
Linting Xue, Noah Constant, Adam Roberts, Mihir Kale, Rami Al-Rfou, Aditya Siddhant, Aditya Barua, and Colin Raffel. 2021.
\newblock \href {https://aclanthology.org/2021.naacl-main.41} {m{T}5: A massively multilingual pre-trained text-to-text transformer}.
\newblock In \emph{Proceedings of the 2021 Conference of the North American Chapter of the Association for Computational Linguistics: Human Language Technologies}, pages 483--498.

\bibitem[{Zhong et~al.(2022)Zhong, Ding, Zhan, Qiao, Wen, Shen, Liu, Yu, Du, Chen et~al.}]{Zhong:2022:Vega}
Qihuang Zhong, Liang Ding, Yibing Zhan, Yu~Qiao, Yonggang Wen, Li~Shen, Juhua Liu, Baosheng Yu, Bo~Du, Yixin Chen, et~al. 2022.
\newblock Toward efficient language model pretraining and downstream adaptation via self-evolution: A case study on {SuperGLUE}.
\newblock \emph{arXiv preprint arXiv:2212.01853}.

\end{thebibliography}

\section{Language Resource References}

\bibliographystylelanguageresource{lrec-coling2024-natbib}
\bibliographylanguageresource{language_resources}

\end{document}